\def\year{2021}\relax
\documentclass[letterpaper]{article} 
\usepackage{aaai21}  
\usepackage{times}  
\usepackage{helvet} 
\usepackage{courier}  
\usepackage[hyphens]{url}  
\usepackage{graphicx} 
\usepackage{amsmath}
\usepackage{tabularx}
\usepackage{bm}
\usepackage{soul}
\usepackage{textcomp}
\usepackage[most]{tcolorbox}
\usepackage{caption}
\usepackage{natbib}  
\usepackage{caption} 
\DeclareMathOperator{\bert}{BERT}
\DeclareMathOperator{\du}{DU}
\DeclareMathOperator{\edu}{EDU}
\DeclareMathOperator{\start}{START}
\DeclareMathOperator{\terminal}{STOP}
\DeclareMathOperator{\mlp}{MLP}

\title{Deep Discourse Analysis for Generating Personalized Feedback in Intelligent Tutor Systems}

\author {
        Matt Grenander,\textsuperscript{\rm 1}
        Robert Belfer,\textsuperscript{\rm 2}
        Ekaterina Kochmar,\textsuperscript{\rm 2,3}
        Iulian V. Serban,\textsuperscript{\rm 2} \\
        François St-Hilaire,\textsuperscript{\rm 2}
        Jackie C. K. Cheung\textsuperscript{\rm 4,5} \\
}
\affiliations {
    \textsuperscript{\rm 1} University of Edinburgh \\
    \textsuperscript{\rm 2} Korbit Technologies \\
    \textsuperscript{\rm 3} University of Cambridge \\
    \textsuperscript{\rm 4} McGill University \\
    \textsuperscript{\rm 5} Mila \\
    matt.grenander@ed.ac.uk, \{robert, ekaterina, iulian, francois\}@korbit.ai, jcheung@cs.mcgill.ca
}

\begin{document}
\maketitle

\begin{center}
\begin{minipage}[t]{\columnwidth}
\vskip3pt
\begin{abstract}
We explore creating automated, personalized feedback in an intelligent tutoring system (ITS). Our goal is to pinpoint correct and incorrect concepts in student answers in order to achieve better student learning gains. Although automatic methods for providing personalized feedback exist, they do not explicitly inform students about which concepts in their answers are correct or incorrect. Our approach involves decomposing students answers using neural discourse segmentation and classification techniques. This decomposition yields a relational graph over all discourse units covered by the reference solutions and student answers. We use this inferred relational graph structure and a neural classifier to match student answers with reference solutions and generate personalized feedback. Although the process is completely automated and data-driven, the personalized feedback generated is highly contextual, domain-aware and effectively targets each student's misconceptions and knowledge gaps. We test our method in a dialogue-based ITS and demonstrate that our approach results in high-quality feedback and significantly improved student learning gains.
\end{abstract}
\end{minipage}
\end{center}

\section{Introduction}
Intelligent tutoring systems (ITS) are AI-powered computer systems that provide personalized tutoring without human teachers \cite{autotutor}. ITSs have been shown to be more effective teachers than other computer tutoring methods and conventional classroom learning \cite{kulik2016effectiveness}. Dialogue-based ITSs use natural language conversations and are particularly appealing as they closely simulate natural human-teacher interactions.

Developing dialogue-based ITSs face many challenges due to the difficulty of parsing and responding to natural text. One critical issue is their ability to provide effective automated feedback. Conventional tutoring typically includes direct guidance from a human teacher, and we would like to incorporate this teaching aspect into ITSs. In fact, previous work has shown that personalized feedback can considerably improve student learning outcomes in an ITS \cite{hint_generation}.

\begin{minipage}[t]{\columnwidth}
\vskip1pt
\hskip-17pt
\centering
\begin{tabularx}{\columnwidth}{|X|}
    \hline
    \textit{System}: You are given a dataset of images of wildlife in Africa. You are tasked with building a model which can identify animals in the images. Is this a regression or classification problem? Try to explain why. \\
    \textit{Student}: I think it's a classification task.  \\
    \hspace{-0.15cm}\begin{tcolorbox}[colback=white,colframe=red,left=0cm,top=0cm,bottom=0cm,right=0cm,before={\vspace{-0.2cm}},]
    \textit{System}: `it's a classification task' is correct. Try supplying a reason for this idea.
    \end{tcolorbox}
    \textit{Student}: It's a classification task, because each animal is a discrete class. \\
    \textit{System}: That's correct! \\
    \hline
\end{tabularx}
\label{tab:ex}
\end{minipage}\hfill
\begin{minipage}[t]{\columnwidth}
\vskip8pt
\captionof{table}{An example interaction with the Korbit ITS, with our feedback mechanism in the red box. The feedback system identifies which concepts are correct and incorrect in the student answer attempt, then provides guidance to fix their answer. In this case, the student has forgotten to justify their answer and is prompted for the missing concept.}
\end{minipage}

\vspace{5pt}In this paper, we present a novel, automated feedback system based on deep learning approaches to discourse. 
Deep learning-based discourse models have produced successful results in recent years across several discourse tasks, such as segmentation \cite{wang-etal-2018-toward}, classification \cite{dai-huang-2018-improving}, parsing \cite{mabona-etal-2019-neural} and tagging \cite{dasigi2017experiment}. 
These successes encourage us to explore applying neural discourse techniques in an ITS. 

Our model takes advantage of established discourse frameworks such as Rhetorical Structure Theory \cite{Mann1988RhetoricalST} to break apart student solution attempts and analyze individual concepts within the student answer.
After the analysis, the system personalizes its feedback to the student attempt and directly indicates which parts are correct, incorrect or missing.
An example interaction with the feedback system is shown in Table \ref{tab:ex}.

Our method extends beyond simply applying existing discourse analysis techniques, since these methods cannot be used on their own for providing feedback in ITSs.
In particular, we introduce a new structure called an exercise graph. 
We construct this graph using discourse components from reference solutions and previous student solution attempts for each exercise in the ITS.
Each graph represents a general solution to the problem, and the model uses them extensively to determine the best feedback.

Our approach contrasts considerably with existing feedback models, such as \citet{deep-tutor} and \citet{hint_generation}.
While these systems can capably guide students towards fixing incorrect ideas, they cannot directly comment on which concepts the students have misunderstood. 
In this way, our system is a novel, practical approach for automatically providing feedback in dialogue-based ITSs. 


We test our method using the Korbit ITS,\footnote{\url{https://www.korbit.ai}} a large-scale, mixed-interface, dialogue-based ITS. In Korbit, students watch video lectures on STEM topics, then answer exercises on the topic. Students' attempted solutions are matched against numerous reference solutions using a machine learning algorithm. We compare student learning gains on Korbit using our feedback mechanism against both simple and strong baselines, and demonstrate that the feedback model results in the highest learning gains and feedback quality.

Our main contribution is an original, personalized feedback mechanism for dialogue-based ITSs able to directly indicate correct and incorrect concepts to students. 
Moreover, our model surpasses previous state-of-the-art benchmarks on discourse segmentation and classification tasks.

\section{Feature Extraction}\label{sec:fe}
In order to reason about concepts within a student answer, the model first fragments their solution and extracts various discourse features.
It decomposes the student's answer into segments called \textbf{EDUs} (described below), computes a \textbf{semantic embedding} for each segment and predicts a \textbf{discourse relation} for each pair of adjacent EDUs.

\subsection{Segmentation}
The segmenter's goal is to divide student input text into distinctive concepts. We follow the Rhetorical Structure Theory (RST) framework for modelling discourse structure \cite{Mann1988RhetoricalST} and use the RST Discourse
Treebank (RST-DT) as our training data \cite{carlson-etal-2001-building}. In RST, text is segmented into non-overlapping units called \textit{elementary discourse units} (EDUs), each of which represents a basic unit of discourse. 

We model discourse segmentation as a sequence classification task, where each word is assigned $1$ if it denotes the start of an EDU and $0$ otherwise. The input text $s$ is first encoded with RoBERTa \cite{roberta} into $m$ embeddings where $m$ is the number of subword tokens in the input:\footnote{See \citet{roberta} and \citet{Wolf2019HuggingFacesTS} for full implementation details.}
\begin{equation}
    h_1,\dots,h_m = \bert(s)
\end{equation}
A binary logistic classifier then classifies each encoding $h_i$:
\begin{equation}
    \hat{y}_i = \sigma(w_s^T h_i + b_s)
\end{equation}
where $w_s$ and $b_s$ are learnable parameters. 

\begin{figure}[t]
    \centering
    \includegraphics[width=\columnwidth]{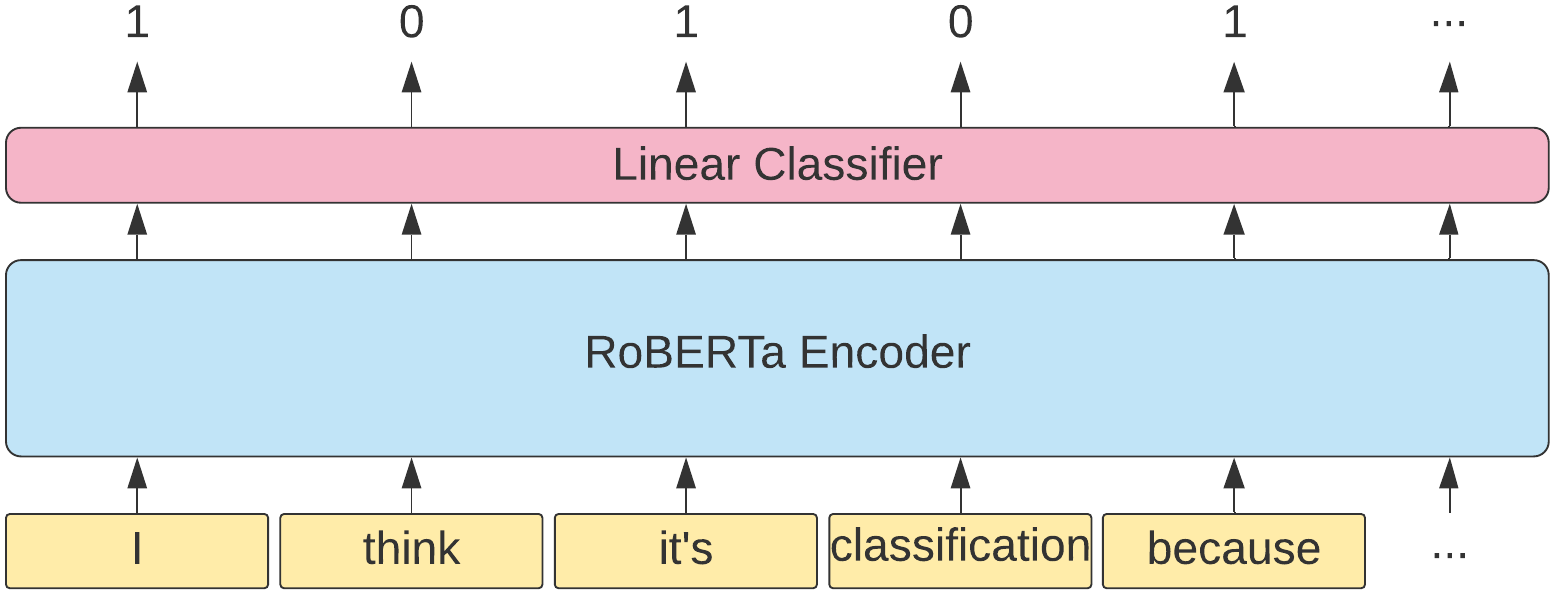}
    \caption{Our neural discourse segmentation model. The input text is tokenized and encoded with RoBERTa. The embeddings are then classified using a linear classifier. Predicting \textit{1} indicates the start of a new EDU, while \textit{0} denotes a continuation.}
    \label{fig:segmentation}
\end{figure}

Figure \ref{fig:segmentation} provides an example of our segmentation approach. For the student input text ``\textit{I think it's a classification task because we are choosing between continuous values}'', the segmenter should predict 1 for tokens \textit{I}, \textit{it's}, and \textit{because}, and 0 for all other tokens. This labelling indicates the correct segments are ``\textit{I think}'', ``\textit{it's a classification task}'', and ``\textit{because we are choosing between continuous values}''.

\begin{figure}[t]
    \centering
    \includegraphics[width=\columnwidth]{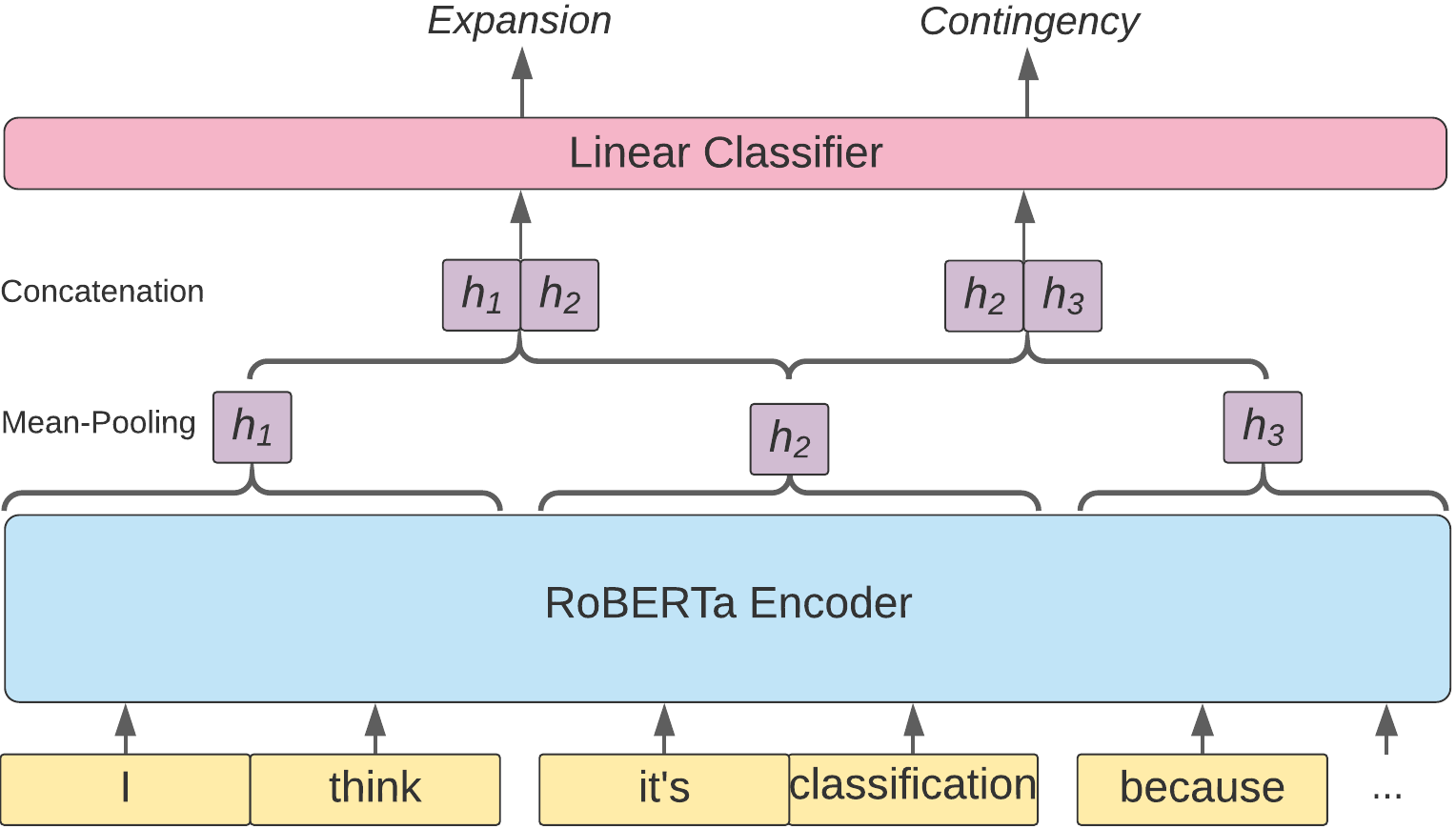}
    \caption{Our neural discourse classification model. After segmentation, the EDUs are encoded with a RoBERTa model, and the output embeddings within each EDU are averaged. 
    These embeddings are then concatenated with adjacent embeddings and classified using a linear classifier.
    Note that the RoBERTa and linear classifier used here are separate from the ones used in segmentation.}
    \label{fig:discourse_classifier}
\end{figure}

\subsection{Semantic Equivalence}
Providing feedback requires understanding which EDUs are semantically equivalent, so the model can match student EDUs to reference solution EDUs and reason about whether these concepts are correct or not.
We use the open-source Sentence Transformers library (SBERT) from \citet{reimers-gurevych-2019-sentence}.\footnote{\url{https://github.com/UKPLab/sentence-transformers}} 
The model is based on fine-tuning BERT-like models on semantic textual similarity (STS) \cite{Cer_2017} and natural language inference (NLI) data \cite{williams-etal-2018-broad, bowman-etal-2015-large}. 
Given raw text, SBERT produces a vector representing semantic content.
This vector can then be compared to other SBERT outputs using their cosine similarity.
A high similarity score indicates the two texts are highly semantically related, while low or negative scores imply orthogonal or opposite meaning.
Although the library is designed for sentences, we find that it works equally well for EDU sentence fragments.
We use SBERT as pretrained by the authors and do not train the model further on our own data.

\subsection{Discourse Classification}
Our approach to feedback generation involves modelling the flow of ideas among typical reference answers, and suggesting which types of ideas are missing or incorrect. Accordingly, the model needs to understand the discourse relations between adjacent EDUs. We therefore train a classifier to predict such relations.

We follow relations defined in the PDTB 2.0 dataset \cite{prasad-etal-2008-penn}, which groups discourse relations into four top-level categories: \textit{Temporal}, \textit{Contingency}, \textit{Comparison} and \textit{Expansion}. PDTB 2.0 defines a classification task in which pairs of discourse units are labelled with the discourse relation between them. The PDTB annotation guide describes discourse units more indistinctly than EDUs in RST, but in practice they are often similar.

PDTB 2.0 contains \textit{pairs} of discourse units, which do not closely resemble the arbitrarily long sequences returned by our neural segmenter model.
We fix this mismatch by extracting paragraphs from PDTB containing consecutive discourse units, using the scripts from \citet{dai-huang-2018-improving}.\footnote{\url{https://github.com/ZeyuDai/paragraph-level_implicit_discourse_relation_classification}} This format better resembles the sequences of EDUs in our data and is more suitable for our purposes.

Our discourse classifier model is similar to \citet{dai-huang-2018-improving}, though we replace their LSTM encoders with the RoBERTa encoder. 
Given a sequence of discourse units $(\du_1,\dots,\du_L)$, we encode each discourse unit using RoBERTa, followed by mean-pooling:

\begin{align}
    h_{i,1},\dots,h_{i,\vert \du_i \vert} &= \bert(\du_i) \\
    \overline{h}_i &= \frac{1}{\vert \du_i \vert} \sum_{j=1}^{\vert \du_i \vert} h_{i,j},
\end{align}
where $\vert \du_i \vert$ is the number of tokens in discourse unit $i$.

The embeddings of each adjacent pair of discourse units are then concatenated and classified into relation categories using a logistic decoder.

\begin{equation}
    \hat{y}_i = \sigma\left(w_d^T \left[\ \overline{h}_i, \overline{h}_{i+1} \right] + b_d \right)
\end{equation}
where $w_d$ and $b_d$ are learnable parameters. Figure \ref{fig:discourse_classifier} illustrates this process.

PDTB 2.0 classifies relations as being either \textit{implicit} or \textit{explicit}, depending on whether there is a cue word such as \textit{``because''} or \textit{``while''}. Following \citet{dai-huang-2018-improving}, we use separate decoder parameters for implicit and explicit cases.
At inference time, since we do not have access to the PDTB explicit/implicit labels, we first scan the discourse units for cue words before applying the relevant decoder.\footnote{The cue words are listed at \url{https://github.com/korbit-ai/deep-discourse-feedback}.}

\subsubsection{PDTB -- RST-DT Mismatch}
Since the segmenter and discourse classifier models are trained using different discourse frameworks (RST-DT vs. PDTB), the two models will produce incompatible features. RST-DT-style EDUs are generally more fine-grained than in PDTB, and certain discourse features are not present in PDTB. For example, the sentence ``\textit{It uses a logistic function to model a binary dependent variable}'' would not be segmented at all in PDTB, but would contain EDUs ``\textit{It uses a logistic function}'' and ``\textit{to model a binary dependent variable}'' in RST-DT.

We reconcile this data mismatch by fine-tuning the discourse classifier on RST-DT-style segments.
We create a dataset with RST-DT-style segments and PDTB-style relations in several steps.
First, we segment the PDTB development and test set with the RST-DT-trained segmenter. We then label the EDU boundaries as follows:
\begin{enumerate}
    \item We map any relations corresponding to shared EDU boundaries, as many EDU boundaries are shared between the two frameworks.
    \item We use cue word heuristics to automatically label EDU boundaries. For example, if we detect ``\textit{if}'' or ``\textit{because}'' near the EDU boundary, we label it as \textit{Contingency}, or if we observe ``\textit{to}'' we annotate it as \textit{Expansion}.
    \item We manually annotate any remaining unlabelled EDU boundaries.
\end{enumerate}
In total, 40\% of relations arise from shared EDU boundaries, 50\% from cue word heuristics and 10\% are labelled manually. 
After first training the discourse classifier on PDTB, we fine-tune it on the generated dataset. 

We assess the fine-tuned model on 103 hand-annotated student solution attempt samples. We find that before fine-tuning on RST-DT-style EDUs, the model reaches 65.7\% accuracy. After fine-tuning, the resulting classifier achieves 81.1\% accuracy, indicating that it can accurately predict relations between RST-DT-style EDUs.

\section{Feedback Model}
Using the features from Section \ref{sec:fe}, we can identify student answer deficiencies and suggest improvements. For each exercise in the ITS, we construct and store a relational graph using reference solutions and all solution attempts across previous students. Each exercise graph represents what a standard solution should resemble for that exercise (Figure \ref{fig:graph}). We train a classifier to recognize which graph transitions are valid. Then, at inference time, we use a local search algorithm in conjunction with the classifier in order to determine what edits, if any, can transform the student’s solution attempt into a better one. Finally, the model uses these results to generate feedback for the student.

\subsection{Graph}\label{sec:graph}
\begin{figure}[t]
    \centering
    \includegraphics[width=\columnwidth]{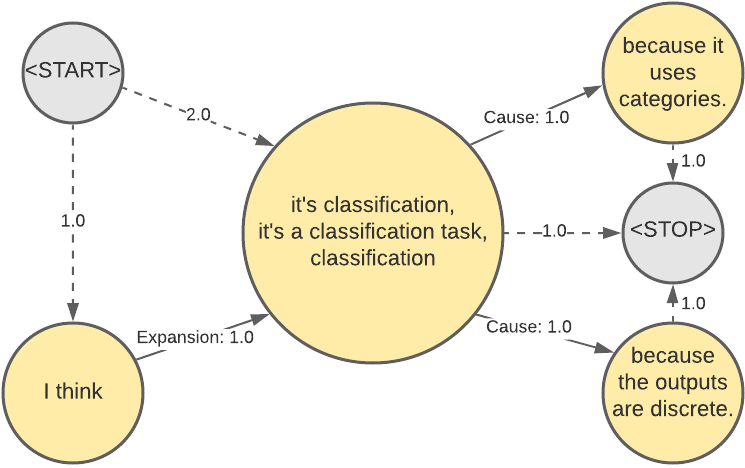}
    \caption{In this exercise graph example, the input texts are ``\textit{I think it's classification}'', ``\textit{it's a classification task, because it uses categories}'', and  ``\textit{classification because the outputs are discrete.}''
    After the EDUs, semantic embeddings and discourse relations are extracted, the semantic embeddings are clustered, and weighted edges added as described in Section \ref{sec:graph}.
    Larger clusters represent commonly occurring concepts, while higher edge weights denote frequently occurring discourse relations.}
    \label{fig:graph}
\end{figure}

The Korbit ITS allows us to collect previous student solution attempts, and we can combine these with the existing reference solutions to construct a graph for each exercise, called an \textbf{exercise graph}.

We first parse all stored student and reference solutions into EDUs, semantic embeddings and relations.
We use DBSCAN \cite{ester1996density} to then cluster the semantic embeddings within each exercise.
For each exercise, we therefore have a set of clusters ${C_1,\dots,C_D}$, where $D$ is the number of clusters determined by DBSCAN. 
Note that $D$ varies depending on the exercise.
Within each cluster, we also track which EDUs correspond to reference or student solutions. Each cluster represents a distinct concept in the exercise, though in practice clustering mistakes may lead to a cluster containing more than one concept, or the same concept mapping to two clusters. 
Larger clusters, especially ones with reference solution EDUs, correspond to more important ideas, while smaller and outlier clusters contain uncommon errors and noise. 
Each cluster represents one node in the graph.

Next, we add weighted edges between the nodes.
For a given reference or student solution attempt, we denote its EDUs as $\{\edu_1,\dots,\edu_L\}$, and the associated clusters as $\{C_{\edu_1},\dots,C_{\edu_L}\}\subseteq\{C_1,\dots,C_D\}$.
We create directed edges $(C_{\edu_i}, C_{\edu_{i+1}})$ for all $i \in \{1,\dots,L\}$, i.e. a directed edge from each EDU's associated cluster to the following EDU's cluster.
We repeat this procedure for all reference and student solutions.
For an edge $(C_s, C_t)$, we set the edge weight equal to the number of adjacent EDU pairs that map to clusters $C_s$ and $C_t$.
Furthermore, edges are subdivided by the relation type predicted by the discourse classifier, meaning that each cluster pair may have up to four weighted edges between them, one for each relation type. 
See Figure \ref{fig:graph} for an illustration of an exercise graph.

The edge weights and relation type determine the importance and discourse relation between two clusters. For example, if two large clusters are primarily joined by a cause relation, then we may expect that the student's answer should contain these two EDUs linked by a causal relation.

In a cluster, if the majority of EDUs correspond to the first EDU in the original text (i.e. $\edu_1$), we designate the cluster as a \textbf{start node}. 
Similarly, if over half the EDUs correspond to the last EDU in the original text (i.e. $\edu_L$), we mark the cluster as a \textbf{terminal node}. 
We can therefore infer whether an archetypal solution should start or end at a particular node.
Note that nodes may be both start and terminal nodes, e.g. if the typical solution consists of a single EDU.

\subsection{Triplet Classifier}
Although the reference solutions often cover the most likely correct solutions, there are often other valid answers not present in the set of reference solutions, which therefore may be non-existent in the graph.
For this reason, rather than working directly with the graphs, we train a classifier to recognize valid and invalid graph transitions.
Following similar prior work \cite{transe}, we refer to this model as a triplet classifier, although in our case the model classifies EDU pairs.
The classifier's aim is to generalize beyond the existing observed graph transitions and learn to infer plausible transitions as well.
This approach allows the feedback mechanism to recognize not only reference answers, but other correct solutions outside of the reference solution set.
There are two stages to training the model: \textbf{data generation} and the \textbf{model architecture}.

\subsubsection{Data Generation}
Training the triplet classifier requires labelled data representing each graph. For each exercise, we create positive samples representing valid transitions in the graph and negative samples for invalid ones.

For a cluster $C$, we use $E_C$ to describe a randomly chosen EDU from $C$. We create positive samples as follows:
\begin{enumerate}
    \item A random cluster $C$ is sampled from the graph, with a sampling probability proportional to the number of EDUs it contains. 
    \item We randomly sample a cluster $C'$ from $C$'s successors, weighted by $C$'s outgoing edge weights. We designate the tuple $(E_C, E_{C'})$ as a positive sample.
\end{enumerate}

Negative samples are constructed similarly:
\begin{enumerate}
    \item A random cluster $C$ is sampled from the graph, exactly as in the positive case.
    \item We sample a second cluster $C''$, excluding the direct successors of $C$. $C''$ must also contain an incoming edge that matches a random outgoing edge from $C$. This matching ensures that we are not simply training the model to recognize incoherent relations, but rather plausibly true, incorrect graph transitions. In 20\% of cases, we instead select a random $C''$, to ensure sample diversity. The tuple $(E_C, E_{C''})$ is labelled negative.
\end{enumerate}

\begin{figure*}
\includegraphics[width=\textwidth]{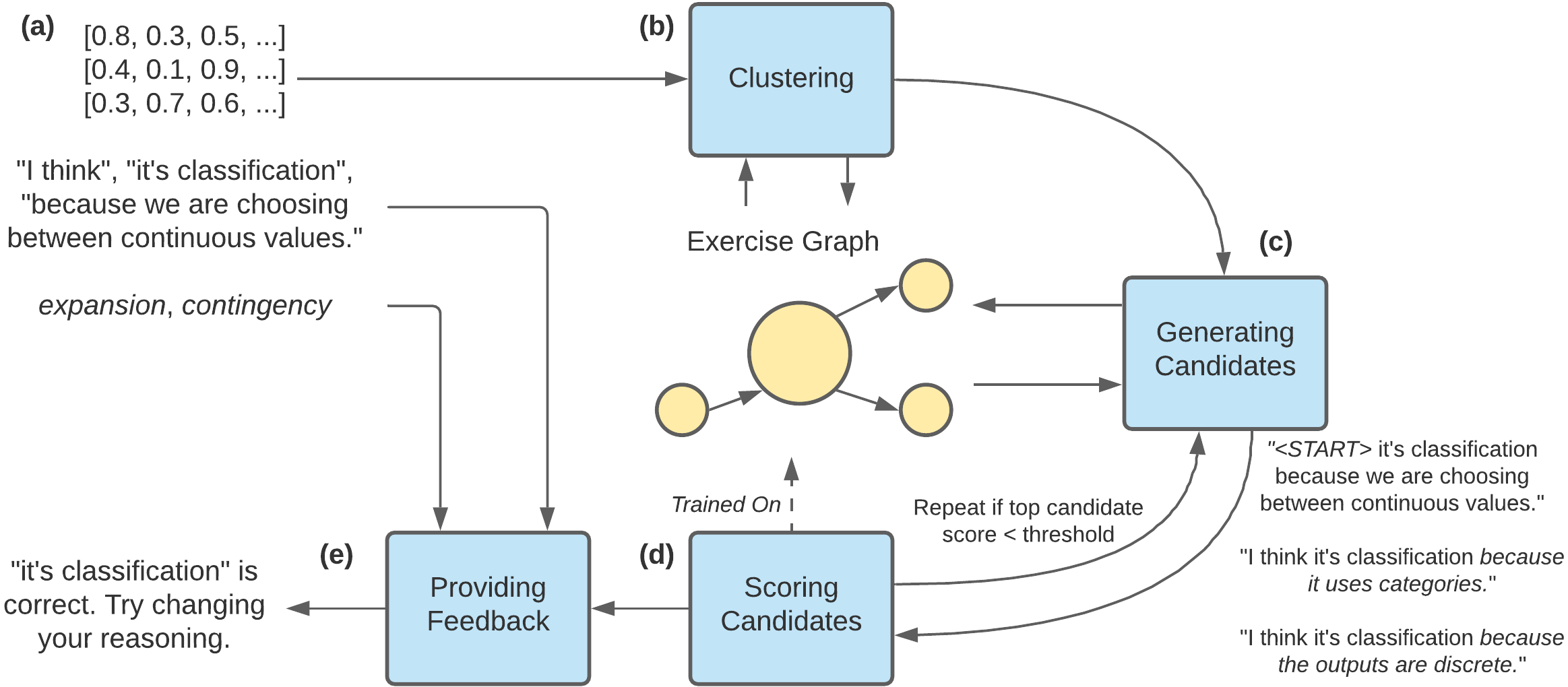}
\caption{An overview of our feedback system after the extraction steps: (a) Features are extracted using the methods described in Section \ref{sec:fe}. From top to bottom, these features include the semantic embeddings from SBERT, the segmented EDUs, and discourse relations between adjacent EDUs. (b) The semantic embeddings are clustered according to the relevant exercise graph. (c) Candidate improvements are generated based on which clusters were matched to the student EDUs, and the clusters' neighbours in the graph. (d) Candidates are scored using the triplet classifier, which is trained offline using the graphs. If the top scoring candidate does not surpass a pre-defined threshold, new candidates are generated and the process is repeated. (e) A feedback message is assembled from the top-scoring candidate and previously computed EDU segments / discourse relations.}
\label{fig:feedback_system}
\end{figure*}

In addition, we generate data representing correct start and terminal nodes.
\begin{itemize}
    \item If a given cluster $C$ is a start node, we create a positive sample $(\langle\start\rangle, E_C)$ and a negative sample $(E_{C''}, E_C)$, where $C''$ is randomly sampled from the graph. 
    \item If $C$ is terminal, we create a positive sample $(E_C, \langle\terminal\rangle)$ and a negative sample $(E_C, E_{C''})$ where again $C''$ is sampled randomly from the graph.
\end{itemize}

\subsubsection{Model Architecture}
The model primarily consists of fine-tuning RoBERTa to minimize a cross-entropy loss. Given a sample $(E_1, E_2)$, we encode both EDUs using RoBERTa and average the output embeddings:

\begin{align}
    h_{1i},\dots,h_{L_i, i} &= \bert(E_i), i\in\{1, 2\} \\
    \overline{h}_i &= \sum_{j=1}^{L_i} h_{j,i}
\end{align}

where $L_i$ is the length of $\edu_i$. The two pooled embeddings are then concatenated and the prediction is generated using a two-layer neural network.

\begin{equation}
    \hat{y} = \mlp\left(\left[\overline{h}_1, \overline{h}_2 \right]\right)
\end{equation}\label{eq:mlp}

This model formulation, which we call the Baseline MLP, does not distinguish between distinct exercises. In order to encode this information, we experiment using a one-hot vector to mark the exercise index. This vector $\mathbf{v}$ is concatenated to $\overline{h}_1, \overline{h}_2$ before the MLP stage:

\begin{equation}
    \hat{y} = \mlp\left(\left[\overline{h}_1, \overline{h}_2, \mathbf{v} \right]\right)
\end{equation}\label{eq:one_hot}

We also experimented with using a separate logistic layer for each exercise instead of the one-hot exercise encoding. However, preliminary experiments showed a clear performance degradation, and we discarded this variant in later experiments.

\subsection{Local Search Inference}
Using the previously described components, we create feedback messages using a local search algorithm.
For a given exercise and student solution attempt, the system parses the student's solution attempt, and matches the EDUs to the exercise graph.
It then attempts to modify the student solution attempt into a correct reference solution by iteratively constructing and scoring candidate solutions drawn from the exercise graph.
Based on the edits needed to transform the student answer into a high-quality reference one, the system devises appropriate feedback for the student.

\subsubsection{Clustering}
After the student solution is parsed into EDUs, relations, and semantic embeddings, we map student EDUs $\{E_1,\dots,E_L\}$ to graph clusters $\{C_{E_1},\dots,C_{E_L}\}$ using DBSCAN \cite{ester1996density}. Note that $\{C_{E_1},\dots,C_{E_L}\}\subseteq\{C_{1},\dots,C_{D}\}$ (the full set of clusters for a graph). EDUs that cannot be clustered are marked as outliers.

\subsubsection{Generating Candidates}
Next, we generate candidate solutions based on the student's solution attempt, similar to generating a beam in beam search.
For each cluster $C_i\in\{C_{E_1},\dots,C_{E_L}\}$, we find $C_i$'s neighbours in the graph, and create candidates by interchanging the EDUs adjacent to $E_i$ with $C_i$'s neighbours.
We generate candidates with $\langle\start\rangle$ and $\langle\terminal\rangle$ tokens if $C_i$ is a start or terminal node in the graph.
For example, using the exercise in Figure \ref{fig:graph}, if the student inputs ``I think it's classification'', we generate candidates ``$\langle\start\rangle$ classification'', ``I think it's classification \textit{because it uses categories}'', and ``I think it's classification \textit{because the outputs are discrete}''.

\subsubsection{Scoring Candidates}
The triplet classifier then scores each candidate by averaging the likelihood of its adjacent EDU pairs.
If none of the candidate scores surpass a predetermined acceptance threshold $\alpha$, we generate and score new candidates using the top-ranked candidate until a maximum number of steps has been reached or the top candidate surpasses $\alpha$.

\subsubsection{Providing Feedback}
Using the top-scoring candidate in the \textit{first} scoring iteration, we determine how the student's EDUs were edited. We check four edit types:
\begin{itemize}
    \item \textbf{Missing}: the student solution is missing EDUs when compared to the top candidate. E.g. if the student has forgotten to explain their reasoning, the system may say, ``\textit{Try supplying a reason for your idea.}''.
    \item \textbf{Excess}: the student's solution contained extraneous EDUs not present in the top candidate. In this case, the system may encourage the student to shorten their answer.
    \item \textbf{Correct Relation}: an EDU was changed, but the discourse relations in both solutions are the same. This situation occurs when the student has supplied the expected type of answer, but some aspects are incorrect: for example, the student entered an answer and an explanation, but the explanation is incorrect. This example is illustrated by Figure \ref{fig:feedback_system}.
    \item \textbf{Incorrect Relation}: similar to `Correct Relation' but the discourse relations have changed. In practice, this scenario rarely occurs.
\end{itemize}

Finally, we record student EDUs that were matched to clusters containing reference solution EDUs. These EDUs are highly likely to be correct, so we attach the correct student EDUs to the final feedback message. See Figure \ref{fig:feedback_system} for an end-to-end overview of the Local Search procedure.\footnote{We provide additional examples in the appendix at \url{https://github.com/korbit-ai/deep-discourse-feedback}.}



\section{Experiments}
For all experiments using RoBERTa, we use the pre-trained HuggingFace `roberta-base' implementation \cite{Wolf2019HuggingFacesTS}.\footnote{\url{https://github.com/huggingface/transformers}}
All graphs are created using scikit-learn's DBSCAN implementation \cite{scikit-learn}, with $\varepsilon=0.15$ and $min\_samples=2$.\footnote{\url{https://scikit-learn.org/stable/index.html}} We base these parameter choices on qualitative observations that a lower $\varepsilon$ value tended to yield better cluster quality. We set the acceptance threshold $\alpha=0.95$ and restrict the local search algorithm to at most 2 iterations (i.e. edits) after observing that student solution attempts are often only missing a single EDU.

\subsection{Segmentation}
We train the segmenter model using the RST-DT dataset, which contains 385 Wall Street Journal (WSJ) articles divided into train (347 articles) and test (38 articles) splits \cite{carlson-etal-2001-building}. We compare the model to \citet{wang-etal-2018-toward}, an LSTM-based segmenter, and \citet{lukasik2020text}, a BERT-based model with extra LSTM layers. We use the same development set as \citet{wang-etal-2018-toward} to fine-tune hyperparameters and we report precision, recall and $F_1$ score on the test set.

\subsection{Discourse Classification}
The discourse classifier is trained using PDTB 2.0 \cite{prasad-etal-2008-penn}, a collection of around 36,000 paired sentences and annotated discourse relations across 2,159 WSJ articles. We follow the same train-validation-test splits as in previous work \cite{rutherford-xue-2015-improving}, and use scripts from \citet{dai-huang-2018-improving} to preprocess the dataset into paragraphs.

As in previous work, we evaluate implicit and explicit relations separately, due to the increased difficulty of implicit cases. For both cases, we report the macro-$F_1$ score and overall accuracy. 
We compare our model against the following two discourse classifiers: 
\begin{itemize}
    \item \citet{dai-huang-2018-improving} extract discourse unit pairs from PDTB to create paragraphs of consecutive discourse units. The main difference between our models is their use of LSTM encoders and a final CRF decoder layer. 
    \item \citet{liu-2020-importance} is a recent model based on RoBERTa with additional attention and pooling layers. Their model is not trained on explicit relations, and therefore we only compare implicit $F_1$ and accuracy scores.
\end{itemize}

\subsection{Triplet Classifier}
\subsubsection{Data Generation}
We generate 124,000 samples, split evenly between positive and negative samples. 
Since the data generation procedure may produce duplicates, we cannot split the data trivially to test generalization.
We instead group duplicates together and place each group into a single data split only,  ensuring that the top 3 most frequent answers are placed in the training set.
This ensures that duplicates are not included in multiple datasets and that generalization accuracy can be properly assessed. 
We create 80/10/10 splits using this procedure for training/validation/test sets, and fine-tune hyperparameters on the validation set.
We report test set accuracy, comparing our models against a majority baseline.

\subsubsection{Model}
We initialize the RoBERTa embeddings using the pretrained SBERT weights \cite{reimers-gurevych-2019-sentence}, as semantic equivalence plays a central role in the triplet classifier's task.
At training time, the Korbit ITS holds 253 exercises in total, and the one-hot vector uses this length.
We encode $\langle\start\rangle$ and $\langle\terminal\rangle$ symbols using RoBERTa's $<s>$ and $</s>$ tokens respectively.
The two-layer neural network decoder uses a 200-dimensional hidden size between first and second layers.
We train the model for 2 epochs.

\subsection{Student Learning Gains}
After integrating the feedback models into the Korbit platform, we collect 247 distinct student interactions with the feedback mechanism in order to measure student learning gains.\footnote{We discard 83 interactions in which the student's first solution attempt was labelled correct by a human annotator.}
The student's second attempt after seeing the feedback is manually labelled as \textit{correct} or \textit{incorrect}. Four domain experts provide the labels.
The average learning gain is then measured as the percentage of times the second student attempt is labelled as correct. 

We compare the feedback model against two baselines.
The first is a \textit{Minimal Feedback} baseline, which simply tells students that their solution is incorrect and they should try again. 
We also compare the feedback model against a \textit{Cluster-Based Feedback} model.
After parsing the student's attempted solution into EDUs and semantic embeddings, the Cluster-Based Feedback uses DBSCAN to match the student EDUs to clusters in the graph.
Then, if any student's EDU successfully matches to a cluster containing at least one EDU from a reference solution, the system indicates to the student that their EDU is correct. 
It then advises the student to re-word other parts of their answer or add additional details without specifying further details.

This baseline tests the effectiveness of the semantic clusters without the relational edges, triplet classifier, and local search. While it uses segmentation and semantic information in the student's answer, it cannot take advantage of the full graph structure to prompt students about discourse relations such as justifications or comparisons. For example, for the interaction listed in Table \ref{tab:ex}, the Cluster-Based Feedback would respond ``\textit{`it's a classification task' is correct. Try re-wording the other parts of your answer or adding additional details}''.

Since other feedback mechanisms exist in Korbit \cite{hint_generation}, we record the number of previous student attempts on a given exercise before interacting with our discourse-based feedback system.
We report both the average learning gain before the second attempt on an exercise, as well as the average learning gain across all attempts on an exercise.

We do not compare our system against other feedback mechanisms in Korbit \cite{hint_generation}, since their formulations differ greatly from ours and are largely orthogonal to our approach. For example, \citet{hint_generation} uses Wikipedia as a resource for hints, while ours does not.

\subsubsection{Direct Feedback Comparison}
In order to intrinsically evaluate model's performance, we also compare the \textit{Minimal Feedback}, \textit{Cluster-Based Feedback} and \textit{Full Feedback} outputs directly in a ranking-based comparison.
Each model's feedback message is ranked w.r.t.\@ its helpfulness on 91 incorrect student solution attempts.
Four domain experts rank the generated feedback.
For each model, we report the percentage of cases the model's feedback is the top-ranked one, allowing for ties.



\begin{table}[t]
    \centering
    \begin{tabular}[\columnwidth]{l|l|l|l}
        \hline
        Model	&	P (\%)	&	R (\%)	&	$F_1$ (\%)	\\ \hline
        \citet{wang-etal-2018-toward}	&	92.9	&	95.7	&	94.3	\\
        \citet{lukasik2020text}	&	95.7	&	96.8	&	94.6	\\
        Our Model	&	$\bm{97.6}$	&	$\bm{97.4}$	&	$\bm{97.5}$	\\ \hline
    \end{tabular}
    \caption{Discourse segmentation results on the RST-DT test set. We measure precision (P), recall (R), and $F_1$ score.}
    \label{tab:seg_results}
\end{table}

\begin{table}[t]
    \centering
    \begin{tabular}{l|ll|ll}
    \hline
	&	\multicolumn{2}{c|}{Implicit}			&	\multicolumn{2}{c}{Explicit} \\
    Model	&	$F_1$	&	Acc	&	$F_1$	&	Acc	\\ \hline
    D and H (2018)	&	48.82	&	57.44	&	93.21	&	93.98	\\
    \citet{liu-2020-importance}	&	63.39	&	69.06	&	-	&	-	\\
    Our Model	&	$\bm{64.43}$	&	$\bm{70.46}$	&	$\bm{94.20}$	&	$\bm{95.15}$	\\ \hline
    \end{tabular}
    \caption{Discourse classification results on the PDTB 2.0 test set. We compare the $F_1$ and accuracy (Acc) scores (\%) against \citet{dai-huang-2018-improving} (D and H (2018)) and \citet{liu-2020-importance}. \citet{liu-2020-importance}'s model is not trained to classify explicit discourse relations and we cannot compare performance on these cases.}
    \label{tab:rel_results}
\end{table}

\begin{table}[t]
    \centering
    \begin{tabular}{l|c}
        \hline
        Model	&	Accuracy (\%)	\\ \hline
        Majority Baseline	&	50.39	\\
        Baseline MLP	&	$79.18\pm0.71$*	\\
        + One-Hot Exercise	&	$\bm{81.20\pm0.68}$\textbf{**}	\\ \hline
    \end{tabular}
    \caption{Test set results from generated data representing valid and invalid graph transitions. One-Hot Exercise refers to the one-hot vector indicating exercise index. * indicates significance at the 95\% confidence level over the Majority Baseline, and ** marks significance at the 95\% confidence level over the Baseline MLP.}
    \label{tab:triplet_results}
\end{table}

\begin{table*}[t]
    \centering
    \begin{tabular}{l|cc}
    \hline
	&	\multicolumn{2}{c}{Average Learning Gain (\%)}			\\
    Model	&	All Attempts	&	Before Second Attempt	\\ \hline
    Minimal Feedback Baseline	&	$25.64\pm13.70$	&	$8.33\pm15.64$	\\
    Cluster-Based Feedback      &   $23.53\pm13.25$   & $20.00\pm10.08$ \\
    Full Feedback Model	&	$\bm{51.11\pm14.61}$\textbf{*}	&	$\bm{75.00\pm18.98}$\textbf{*}	\\ \hline
    \end{tabular}
    \caption{Student learning gains on the Korbit ITS. Significant results at 95\% confidence over the Cluster-Based feedback are marked with *.}
    \label{tab:learn_gain_res}
\end{table*}

\section{Results}
\subsection{Segmentation}

The segmentation results for RST-DT are presented in Table \ref{tab:seg_results}.
Our model outperforms \citet{lukasik2020text}'s model, despite using a much simpler model formulation.

\subsection{Discourse Classification}

Discourse classification results on the PDTB 2.0 test set are presented in Table \ref{tab:rel_results}.
Although \citet{dai-huang-2018-improving}'s and our model both rely on chaining adjacent discourse units into paragraphs, using the RoBERTa encoder results in strong implicit relation accuracy gains.
On the explicit cases, accuracy is closer to \citet{dai-huang-2018-improving} but still remains high.
Interestingly, our model achieves similar results to \citet{liu-2020-importance}, although our model design is considerably simpler and extends to explicit relations as well.

\subsection{Triplet Classifier}

The triplet classification results on the generated test set are shown in Table \ref{tab:triplet_results}.
Although there is no previous work for comparison, both models significantly outperform a majority baseline. Using a general $z$-test, we test if the one-hot exercise index vector significantly improves the Baseline MLP. Since the generated test set is so large, we find that the improvements are significant with $p < 0.0001$.

\subsection{Student Learning Gains}

Results from the learning gains experiments are shown in Table \ref{tab:learn_gain_res}. The `All Attempts' column indicates student learning gains across all attempts made, and the `Before Second Attempt' column contains entries in which the student tried only once previously. Our experiments show the Full Feedback Model strongly improves over the baselines in both settings. We test the results for significance using an $N-1\ \chi^2$ test. The Full Feedback Model leads to significant student learning gains compared to the Cluster-Based Feedback in the All Attempts setting with $p=0.0047$ and in the Before Second Attempt setting with $p=0.00006$. 
This result indicates that the graph structure and triplet classifier are crucial to our feedback approach, and the clusters by themselves are not adequate to achieve high-quality feedback.
The inadequacy of the Cluster-Based Feedback is especially apparent in the All Attempts setting, where it does not noticeably improve over the Minimal Feedback Baseline.
We measure inter-annotator agreement using Krippendorff's alpha using 30 samples. We obtain $\alpha=0.831$, which indicates significant inter-annotator agreement.

It is interesting to note that the learning gains are considerably more pronounced when the student has only been given a single hint. Although the result may seem surprising, the difference is most likely due to selection bias of the students. Students who require many hints in order to solve an exercise may face external challenges or have knowledge gaps. 
For example, these students may experience language barriers or lack the prerequisites to solve the problem.

In general, we find that the most common feedback type is the \textit{missing} class, as students often answer only half the question and omit aspects such as explaining their answer. In contrast, the \textit{wrong relation} feedback is rarely used, as generally, students are aware of which relation is expected even if the EDU supplied is incorrect.

\subsubsection{Direct Feedback Comparison}

\begin{table}[t]
    \centering
    \begin{tabular}{l|c}
    \hline
    Model	&	Top Ranked (\%)	\\ \hline
    Minimal Feedback Baseline	&	$24.18\pm8.80$	\\
    Cluster-Based Feedback	&	$32.97\pm9.66$	\\
    Full Feedback Model	&	$\bm{65.93\pm9.73}$\textbf{*}	\\ \hline
    \end{tabular}
    \caption{Model feedback quality in a ranked comparison test. We measure the proportion each model provides the top ranked feedback (\%), with corresponding 95\% confidence intervals. Since we allow for ties, the values do not sum to 100. Significant results over the Cluster-Based feedback are marked with *.}
    \label{tab:fdbk_comp}
\end{table}

Results from the direct feedback comparison are shown in Table \ref{tab:fdbk_comp}. The Full Feedback Model's feedback is ranked best in 65.93\% of cases, substantially higher than other models. The Full Feedback Model results compared to the Cluster-Based Feedback are significant with $p < 0.0001$, using an $N-1\ \chi^2$-test. 
This result parallels the Student Learning Gains experiments, demonstrating the Full Feedback Model provides more meaningful feedback compared to baselines.

We measure inter-annotator agreement by gathering each pair of annotator's labels and computing the Spearman's rank correlation between all pairs. The result is $\rho=0.868$, which indicates a very high agreement between annotators.

\section{Discussion}
The results of both the learning gains and feedback ranking evaluation demonstrate that the Full Feedback Model is an effective feedback tool for ITSs.
The comparison against the Cluster-Based Feedback model further demonstrates the importance of incorporating the rhetorical structure and the inferred relationships into the feedback generation process.

Nevertheless, we did notice cases where the model's feedback was subpar or even nonsensical.
One of the key weaknesses lies with the triplet classifier. Since the classifier's role is critical for selecting the right feedback, its errors are often amplified in the final feedback message. For example, suppose that the correct solution is the EDU sequence $[E_1, E_2]$ and the student's attempt is $[E_1, E_3]$. If the highest scoring candidate is instead $[E_1, \langle\terminal\rangle]$, then the feedback model will mistakenly tell the student to shorten their answer, instead of recommending $E_2$.

Future methods may benefit from softening the classifier's role within the feedback mechanism, or by trying to increase accuracy. One option is to encode the exercise's problem text when learning graph transitions. The current formulation encodes the exercise as a one-hot vector, ignoring the problem statement text.
Richer exercise representations that include the question may lead to better classifier accuracy.
Another potential improvement is to change the triplet classification task into a ranking task, where the model must learn to rank reference solutions over student solution attempts.

Our approach leverages four distinct RoBERTa models, which consume a considerable memory footprint. Each subtask is also optimized separately rather than jointly optimized towards the final feedback generation task.
Having numerous subtasks results in difficulties with maintaining many separate moving parts, as well as computational issues.
We explored combining related tasks such as segmentation and discourse classification in a multi-task setup, but found that performance on both tasks declined substantially. Future work may analyze methods to simplify the feedback procedure without sacrificing accuracy.

Although our experiments only consider STEM exercises, the feedback mechanism is not exclusively tuned towards STEM and could easily be adapted to other subjects. 
We expect that topics requiring less domain-specific knowledge will be easier to adapt, as the semantic equivalence model (SBERT) is not intended for domain-transfer tasks.
In our experiments, SBERT occasionally struggled with domain-specific concepts, e.g. equating the ideas of \textit{bias term} and \textit{intercept} in the context of linear models.

There are many other directions worth exploring. One possibility is to replace the fixed feedback rules with a seq2seq model, which would gather edits from each step of the local search and output feedback based on the edit path.
Future work may also consider generating multiple feedback messages and training a classifier to choose among them.

\section{Related Work}\label{sec:rel_work}

Discourse segmentation is traditionally viewed as a preliminary step in RST parsing \cite{Mann1988RhetoricalST}, though other applications, such as summarization, exist as well \cite{li-etal-2020-composing}. Past work on discourse segmentation includes using SVMs \cite{hernault2010hilda}, LSTMs with CRFs \cite{wang-etal-2018-toward}, and more recently, BERT-based architectures \cite{lukasik2020text}. Discourse segmentation has also been explored in other contexts such as medical \cite{ferracane-etal-2019-news} and multilingual settings \cite{muller-etal-2019-tony}.

The PDTB corpus \cite{prasad-etal-2008-penn} has motivated many studies on discourse classification, especially on the challenging implicit discourse case \cite{pitler-sense, lin-recognizing, rutherford-xue-2015-improving}.
Past approaches have used feature-based statistical models \cite{pitler-sense}, LSTMs with attention \cite{liu-implicit}, and BERT-based architectures \cite{liu-2020-importance}.
Our work is most similar to \citet{dai-huang-2018-improving}, since we follow their approach in organizing discourse units into paragraphs. Our model shares characteristics with \citet{liu-2020-importance}, whom also employ a BERT-based approach to discourse classification.

An alternative to our discourse approach is full RST parsing. Many RST parsers exist \cite{braud-etal-2017-cross-lingual, ji-eisenstein-2014-representation, feng-hirst-2014-linear}, but highly accurate RST parsing remains out of reach due to the complex nature of RST trees. The majority of student answers on Korbit are less than 3 EDUs and therefore building a complex tree structure may be unnecessary.

Previous work has studied representing graph transitions with neural networks \cite{transe, fan-etal-2014-transition}. Local coherence modelling, in which models learn to rank sentences in a logical order \cite{barzilay-lapata-2008-modeling}, is another similar task. 

Many ITSs provide feedback guiding students towards the correct solution in an exercise \cite{itads, deep-tutor, autotutor, ibm_tutor, Nakhal, Aldahdooh, tamura2015generating, guo2016questimator, shah2017automatic, Serban_2020, hint_generation}. However, to our knowledge, none have previously employed a neural discourse-based mechanism for personalized feedback.

\section{Conclusion}
In this work, we investigate how we can use deep learning, relational graphs, and discourse analysis to develop a data-driven, automated, personalized feedback mechanism. Our approach decomposes students' attempted solutions into EDUs and relations, then matches their attempt to a graph representing the exercise. Based on a traversal of this graph, personalized feedback for each student is generated on the fly. Although it is not our primary goal, our discourse segmentation and classification models also attain new state-of-the-art performance on RST-DT and PDTB 2.0, respectively.

Evaluating the feedback quality and student learning gains on Korbit show that our feedback model markedly improves the students' learning experience.
The results suggest that our feedback system can effectively help students and that discourse analysis may be a highly effective approach for building personalized feedback systems.

\section{Acknowledgments}
This work was conducted while the first author was employed at Korbit for an internship.

{\fontsize{9.4pt}{10.4pt} \selectfont \bibliography{bibliography}}

\end{document}